# Fine-Tuned Language Models for Domain-Specific Summarization and Tagging

Authors: Jun Wang, Fuming Lin, Yuyu Chen

ZhejiangLab, Yuhang district, Hangzhou 311121, China

## Abstract

This paper presents a pipeline integrating fine-tuned large language models (LLMs) with named entity recognition (NER) for efficient domain-specific text summarization and tagging. The authors address the challenge posed by rapidly evolving sub-cultural languages and slang, which complicate automated information extraction and law enforcement monitoring. By leveraging the LLaMA Factory framework, the study fine-tunes LLMs on both generalpurpose and custom domain-specific datasets, particularly in the political and security domains. The models are evaluated using BLEU and ROUGE metrics, demonstrating that instruction fine-tuning significantly enhances summarization and tagging accuracy, especially for specialized corpora. Notably, the LLaMA3-8B-Instruct model, despite its initial limitations in Chinese comprehension, outperforms its Chinese-trained counterpart after domainspecific fine-tuning, suggesting that underlying reasoning capabilities can transfer across languages. The pipeline enables concise summaries and structured entity tagging, facilitating rapid document categorization and distribution. This approach proves scalable and adaptable for real-time applications, supporting efficient information management and the ongoing need to capture emerging language trends. The integration of LLMs and NER offers a robust solution for transforming unstructured text into actionable insights, crucial for modern knowledge management and security operations.

## 1.1 Background

Named entity recognition identifies and classifies key information—like names, locations, and organizations—in text. It helps automate information extraction, improves search accuracy, and supports data organization by transforming unstructured text into structured, actionable insights for various applications such as search engines and knowledge management. [1,2]

The rapid evolution of sub-cultural languages and slang creates significant challenges for language communication[3]. These dynamic forms often diverge from traditional dictionary standards, making them difficult to generalize or interpret using conventional linguistic resources. Such linguistic innovation can be exploited by criminals, who use codewords and evolving jargon to bypass automated internet filters. This complicates law enforcement efforts, as tracking and decoding these non-standard expressions becomes increasingly difficult, hindering effective monitoring and intervention. For example: "cosplay", "steampunk", "furry fandom", employ unique terms and references that function almost like "secret languages" to outsiders.

The overwhelming volume of information available today makes it impractical for any individual to read and analyze every detail line by line. However, large language models can efficiently summarize and categorize this data, transforming it into easily recognizable formats. This enables relevant authorities to quickly identify critical insights and focus their attention where needed, thereby enhancing their ability to respond to emerging threats and fight crime more effectively through informed decision-making and targeted intervention.

In recent years, large language models have demonstrated remarkable power and accuracy in reading and understanding vast corpuses of text, enabling them to extract insights and recognize patterns at a scale unattainable by humans[4]. However, a significant challenge arises from the rapid evolution of new slang and sub-cultural vocabulary, which necessitates frequent and targeted fine-tuning of these models to ensure they remain effective at interpreting emerging language trends and accurately capturing the meaning of novel expressions.

In this work, we experimented with various fine-tuning approaches using corpora from specialized fields that employ specific vocabularies, aiming to generate concise summaries that immediately facilitate post-training named entity tagging; by leveraging multiple models available on the Llama factory platform, we evaluated the quality of these summaries using BLEU[5] and ROUGE[6] scores, and found that coupling summary generation with named entity tagging creates an extremely effective system for topic recognition, which in turn enables a powerful and rapid document distribution pipeline, ultimately demonstrating that our approach offers a fast, convenient, and scalable solution for processing domain-specific texts and supporting efficient information management for downstream applications.

### 1.2 Llama Factory

LLaMA Factory is an open-source, unified framework designed to streamline the efficient fine-tuning of over 100 large language models—ranging from LLaMA and Mistral to Qwen and ChatGLM—without requiring extensive code. Leveraging techniques like low-rank adaptation (LoRA), QLoRA, full parameter updates, and reward-based approaches, it enables scalable adaptation across diverse architectures. Users can interact through a zero-code CLI or an intuitive Web UI (LLaMABoard), making it accessible for both technical and non-technical users[7,8].

A key innovation is its prompt engineering pipeline, where users craft or refine prompts to guide the model's behavior during fine-tuning. By integrating specialized domain corpora and carefully designed prompt templates (e.g., instruct-style or completion contexts), LLaMA Factory teaches the model to focus on target tasks—such as summarization or named-entity tagging—with higher precision. These prompts are baked into the YAML configuration and incorporated into training loops via LoRA or QLoRA, enabling lightweight yet effective adaptation.

### 1.3 Named Entity Recognition

Named entity recognition (NER) algorithms typically use statistical models or neural networks to identify and classify entities such as names, locations, and organizations within text by analyzing contextual patterns, word boundaries, and syntactic cues. NER excels at extracting structured information from unstructured data, offering high precision in entity identification even in noisy or domain-specific corpora. While large language models can perform NER as part of broader tasks, dedicated NER algorithms remain advantageous due to their efficiency, interpretability, and ability to generalize well with limited training data, making them especially valuable for targeted information extraction and real-time applications.

**Glossary**

BLEU (Bilingual Evaluation Understudy): a commonly used metric for evaluating machine translation quality. BLEU-4 stands for the four-gram BLEU score, which measures the n-gram match between the model's generated text and the reference text, where n=4. Higher values indicate greater similarity between the generated text and the reference text, with a maximum value of 100.

ROUGE (Recall-Oriented Understudy for Gisting Evaluation) is a metric used to evaluate the performance of automatic summarization and text generation models. ROUGE-1 represents the unigram ROUGE score, and ROUGE-2 represents the bigram ROUGE score. These metrics measure the degree of match between single-word and bigram sequences, respectively, between the generated text and the reference text. Higher values indicate greater similarity between the generated text and the reference text, with a maximum value of 100.

**Methods**

**2.1 Experimental Objective**

This study aims to evaluate the efficacy and accuracy of instruction fine-tuning for large language models (LLMs) in the context of domain-specific data summarization. The primary objective is to quantitatively assess the performance improvement, or degradation, that instruction fine-tuning imparts on models when processing both general-purpose and specialized domain data.

**2.2 Models and Rationale**

Due to the current lack of support for the LLaMa3.1-8B-Chinese-Chat model within the LLaMa-Factory framework, two alternative benchmark models were selected to facilitate a controlled comparison. The models used in this experiment are the LLaMa3-8B-Instruct and the LLaMa-8B-Chinese-Chat. These models serve as baselines to test the accuracy on both general and domain-specific datasets before and after the instruction fine-tuning process is applied.

**2.3 Datasets and Preprocessing**

The evaluation was conducted using two distinct types of data to ensure a comprehensive assessment of model performance.

**General Data Benchmarking:** For the general data evaluation, we utilized two built-in datasets from the LLaMa-Factory library: alpaca and glaive_toolcall. These datasets provide a broad foundation of instructional tasks. A standardized split was employed, with 10% of the combined data from these sets reserved as a held-out test set to evaluate model generalization.

**Domain-Specific Data Fine-tuning:** The core of this investigation centers on a custom, domain-specific dataset provided by Professor Liu. This dataset, comprising 4,905 data points, was meticulously processed and formatted to be compatible with the model's expected input structure. Following preprocessing, the dataset was partitioned into training, validation, and test sets using an 8:1:1 ratio. This resulted in a dedicated test set of 405 data points, ensuring a robust and unbiased evaluation of the fine-tuned model's performance on unseen domain information.

**2.4 Experimental Procedure**

The experimental procedure was consistent for both data types. For each model, we first established a baseline by measuring its accuracy on the respective test sets (general and domain) in their pre-trained, out-of-the-box state. Subsequently, the models underwent instruction fine-tuning on the corresponding training datasets. The performance was then re-evaluated on the same test sets post-fine-tuning. The key metric for comparison was the change in accuracy, which directly indicates the performance impact of the instruction fine-tuning process.

**2.5 Computational Environment and Tools**

All experiments were conducted on a high-performance computing node equipped with an AMD Ryzen 9 7950X 16-Core Processor, 128 GB of system memory, and an NVIDIA RTX A6000 GPU with 48 GB of VRAM. The software stack consisted of Python 3.11.5 and CUDA 12.4. The primary tool used for orchestrating the fine-tuning experiments and evaluations was LLaMa-Factory, which provides a standardized and reproducible pipeline for model training and assessment.

**Data**

Alpaca: The Self-Instruct Alpaca dataset is a pioneering 52K-instruction dataset generated using OpenAI's text-davinci-003. Its purpose was to democratize instruction-following models by providing a cost-effective way to teach LLMs to understand and follow user intent, enabling the creation of small, capable models like Stanford Alpaca without massive human annotation.

Glaive: The Glaive dataset is a larger-scale synthetic dataset (over 1.4 million instructions) generated using a powerful LLM guided by a "skill graph." Its purpose is to enhance LLM reasoning and complex task performance by providing dense, multi-step instructions that cover a wide range of capabilities, pushing models beyond simple Q&A to execute sophisticated chains of thought.

## Analysis

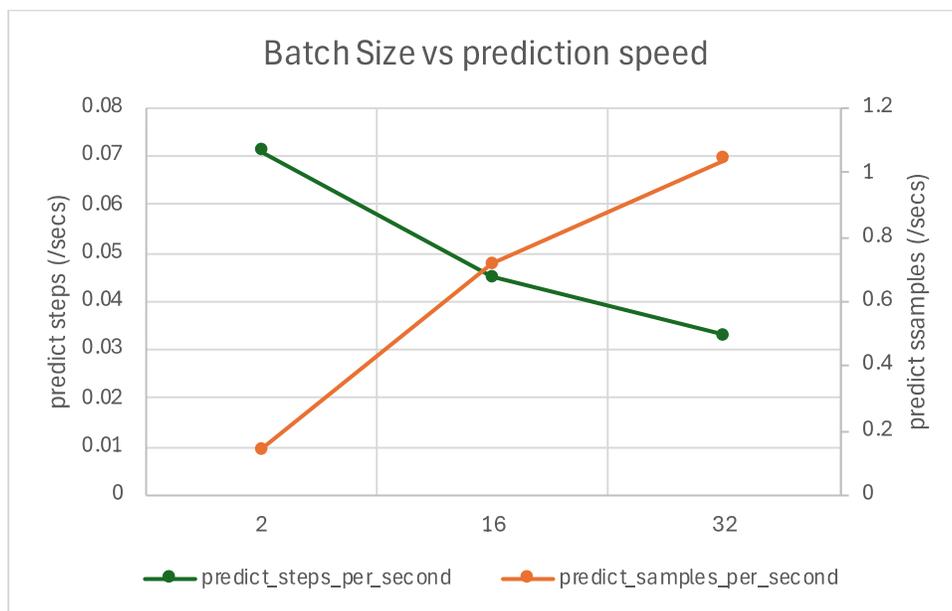

*Figure 1: The time-dependency of prediction steps per second and prediction samples per second over the batch size to the model*

Using the Llama factory, we made an initial quantification on the speed for testing. Prediction samples per second indicates the number of samples a model can generate per second. It is often used to evaluate the inference speed of a model. Prediction steps per second refers to the number of steps the model can execute per second. For generative models, this generally refers to the number of generative operations executed per second. Figure 1 shows that predict samples per second increases linearly with batch size, whilst predict steps per second acts the opposite of decreasing with batch size. As batch size increases, the computational load and memory pressure for each single generative step grow significantly. While larger batches improve overall throughput, the latency per step increases, meaning each step takes longer to complete. This results in fewer steps being executed per second. On the other hand, Increasing batch size improves hardware parallelism and amortizes overhead costs. While each individual batch takes longer to process, the total number of samples processed per second increases significantly. This demonstrates the classic throughput vs. latency trade-off, where one could sacrifice single-operation speed for greater overall efficiency. Henceforth, the batch size is set to 16 throughout the work.

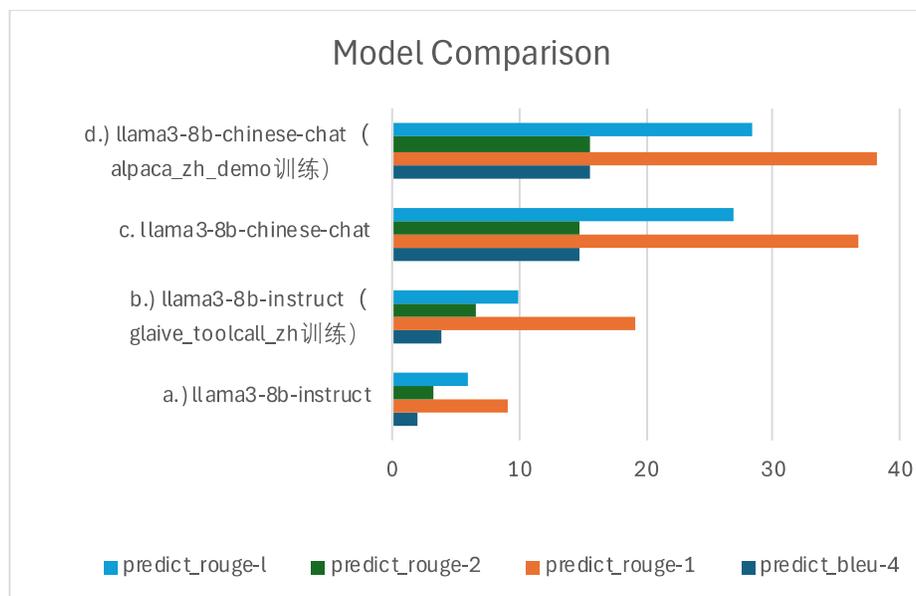

*Figure 2: Performance compairson of the different model versions over the alpaca data*

On Llama factory, we selected the generic Llama3-8b-instruct to act as a baseline, and measured the Bleu score, a commonly used metric for evaluating machine translation quality which measures the n-gram match between the model's generated text and the reference text, and the Recall-Oriented Understudy for Gisting Evaluation (or ROUGE) score, a metric used to evaluate the performance of automatic summarization and text generation models, for the Alpaca test data. Note that Rouge-1, 2 and L refers to unigram, bigram and Longest Common Subsequence, respectively, which corresonds to measures the degree of match between the n-gram and the reference text.

Figure 2 shows the comparison between Llama3-8b-instruct (a), which is trained predominently with English text corpus, and its counterpart Llama3-8b-chinese-chat (c) which is predominantly trained with Chinese text corpus. There is a clear signficant improvement in the summarization performance, as expected on targeted chinese text corpus. This improvement cannot be achieved by fine-tuning a the Llama3-8b-instruct model with chinese text on a post-training stage, see (b). Despite this, further fine-tuning of the Llama3-8b-chinese-chat in the hope to enhance the model's Chinese comprehensibility did not find any significant boost (see c and d) across all the scores.

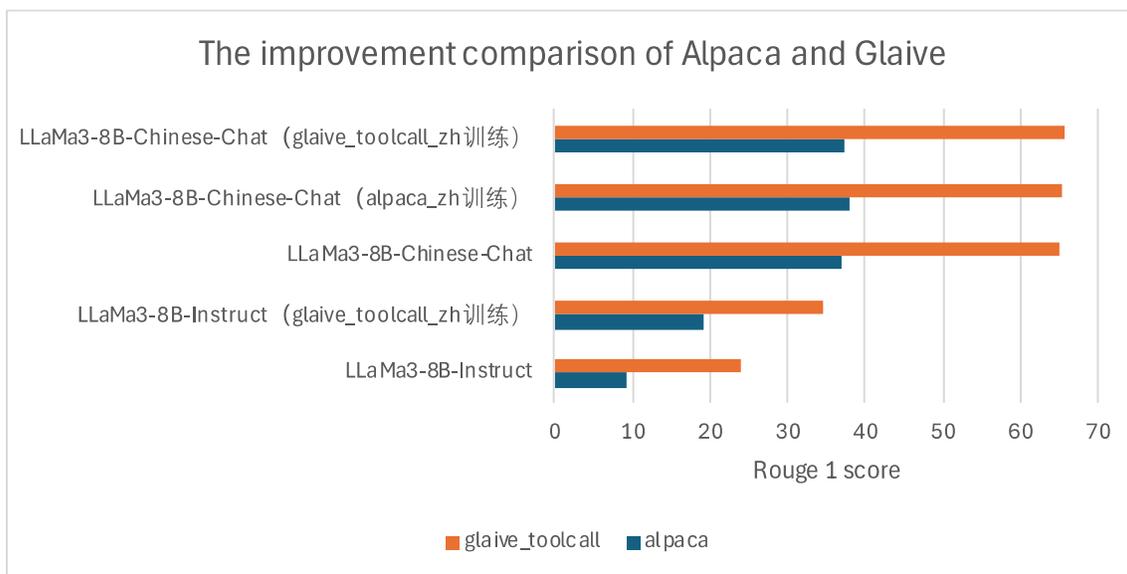

*Figure 3: Comparing the sample datasets of alpaca and Glaive over different models*

Figure 3 shows how the performance changes across two open-source datasets which focuses on Chinese language. As aforementioned, the Llama3-8B-Chinese chat dominates over the LlaMa-8B-Instruct. This is also true even after Llama3-8B instruct is fine-tuned with Chinese language corpuses. It is noteworthy that Glaive has a generally higher performance than Alpaca due across all models because it consists of relatively shorter and less complex questions. Furthermore, it is worth stating the fine-tuned Llama3-8B-Chinese chat models with Alpaca and Glaive data separately sees a minute difference in the performance when answering the questions from the corresponding datasets, i.e. Alpaca is 27.3 to 28.3 and Glaive is 59.1 and 59.3. We expected these intuitive results.

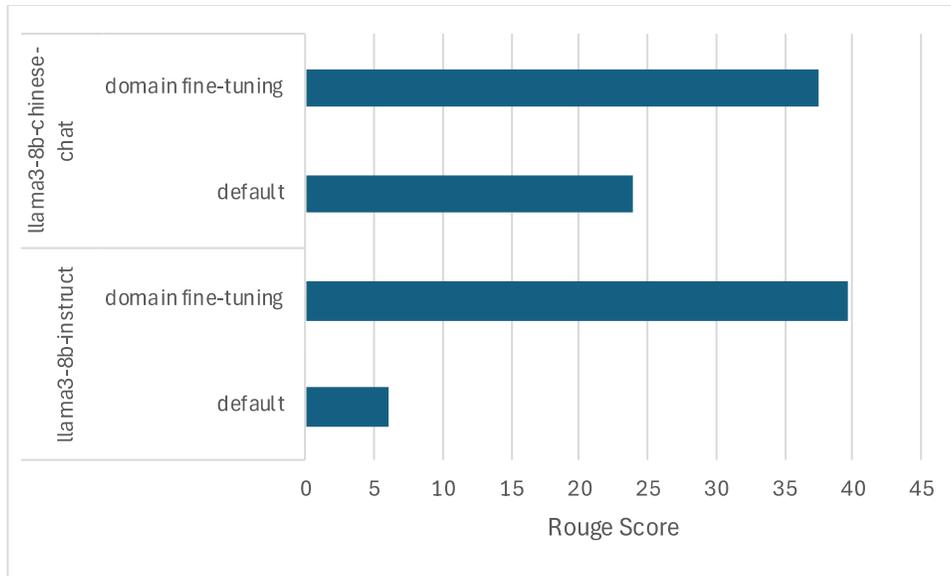

*Figure 4: The performance difference between Llama-instruct and Llama-Chinese with and without domain fine-tuning enhancement*

Figure 4 shows the use of Llama3 shows that for domain data, instruction fine-tuning can greatly improve the prediction accuracy. This applies to both the Llama3-8b-instruct, from 6.0 to 39.7 and the Llama3-b-chinese-chat model from 24.0 to 37.5. The surprising result is the former model outperformance the latter model after fine-tuning. The same pattern is also seen on the BLEU score metrics. This is possible if the former model that is predominantly trained with English corpora, especially those from the high-quality web-scraped data used for major models, contain a vast amount of structured knowledge, scientific papers, code, and logical reasoning examples. If Llama3-8b-instruct was trained on a higher-quality, more diverse dataset that built stronger reasoning skills, it can transfer these capabilities to the Chinese task. It's not just about language, but the underlying intelligence. A "smarter" model can learn to apply its intelligence in a new language more effectively than a model that is fluent but less capable.

*Figure 5: Demonstration of the named entity tagging after the model summariser*

Figure 5 shows the result of a test example input document to that is summarized by the Llama3-b-chinese-chat model fine-tuned by the domain dataset. The input and output are both entity tagged by the domain dataset. The input and output are both entity tagged by open-source libraries (i.e. Spacy Model English -en_core_web_sm(v3.5.0) and Chinese – zh_core_web_sm (v3.5.0)). The sub-figure (a) shows a longform text tagged with location, organisation, concept and more. It can be seen that many concepts are involved in this descriptive document. Some tagging could also be ambiguous, for example: the term "电力系统" or "power systems" was mis-tagged as an organization, but the context likely refers to the more generic noun. The output condenses the text into a summary, and as a result the tagging is much more condense and still carry the essential concepts of the entire paragraph. Sub-figure (b) shows the translated English version, and it can be seen that the word "researchers" has a PERSON tag whilst the Chinese version has not. This is normal as taggers are lanaguage dependent and may not be consistent after translation, however, the English summary correctly and consistently reflects the same summary of the Chinese text.

## Discussion

Given that NER tagging process usually has an execution completion time in less than a second for short-form texts. We paid more attention on the execution of large-language model. We decided on a batch size of 16 because the prediction steps is approximately 0.05 per second and prediction samples of 0.8 per second is an adequate throughput vs. latency trade-off point. Therefore, integrating this stage with the NER process would give a execution time that is adequate for real-time analysis.

The default model of Llama3-8b-instruct doesn't have a strong Chinese language comprehension than Llama3-8bchinese-chat, which is reflected by various test parameters such as BLEU score and Rouge1, Rouge2 and RougeL scores. This is as expected because both models are trained with very different corpuses. The fine-tuning clearly shows a significantly improvement to the model performance, and is an indespensible process in order to have an acceptable generative capability for question and answer, and summarisation. We have showed this with three datasets, the Alpaca, Glaive and a custom-built domain-specific dataset. The continous fine-tuning is an important process for capturing the new generation of sub-cultural words and concepts in oder to upkeep the a large language model's comprehension to the ongoing fast-evolving subcultural modern languages. Despite that LlaMa-8B-Instruct, in its default pre-trained state delivered very poor performance on the Chinese question and answering tasks of Alpaca and Glaive, we observed that after fine-tuning it with domain-specific datasets, it's performance does not necessarily lack behind the similarly fine-tuned Llama3-8B-Chinese-chat model. We attribute this to the possibility that LlaMa-8B-Instruct was pre-trained with data where logical reasoning examples gave the model a superior intelligence, and fine-tuning unravels its true potential on tackling the question and answering after adapting to the Chinese language. Finally, the NER process is connected to the outout of the Llama3-8B-Chinese-chat model fine-tuned with domain data that was purposely built for converting texts where topics are focused on the political and security domains to concise one-sentence summaries. The result shows a good concise summary that is tagged with topic categories that makes easy for quick identifcation of the document context. This pipeline conveniently integrates the intelligence of large language model with underlying statistical models in the NER algorithm to bring about structured information.

## Conclusion

In this work, we have integrated two components, a large language model and NER process, into a pipeline for outputting tags of a summary from a text of domain-specific topics. We showed that fine-tuning boosted the performance of language comprehension and reasoning. The language model is able to condense a long-form text into a summary without losing the core concepts, and a subsequent NER process structuralize the summary, which is valuable for efficient document topic distribution that could be adapted for capturing conceps even in the rapidly evolving sub-cultural modern languages.

Computational Linguistics (ACL '02). Association for Computational Linguistics, 311–318. https://doi.org/10.3115/1073083.1073135

6. Chin-Yew Lin. 2004. ROUGE: A Package for Automatic Evaluation of Summaries. In Proceedings of the Workshop on Text Summarization Branches Out (TextSum '04). Association for Computational Linguistics, 74–81.

7. Hiyouga. 2024. LLaMA-Factory. GitHub repository. Retrieved June 2024 from https://github.com/hiyouga/LLaMA-Factory

8. Yaowei Zheng, Richong Zhang, Junhao Zhang, Yanhan Ye, Zheyan Luo, Zhangchi Feng, and Yongqiang Ma. 2024. LlamaFactory: Unified Efficient Fine-Tuning of 100+ Language Models. In Proceedings of the 62nd Annual Meeting of the Association for Computational Linguistics (Volume 3: System Demonstrations). Association for Computational Linguistics, 455– 465. http://arxiv.org/abs/2403.13372